\documentclass[a4paper]{article}
\usepackage{iclr2019_conference}
\usepackage[english]{babel}
\usepackage[T1]{fontenc}
\usepackage{caption}
\usepackage[parfill]{parskip}

\usepackage[a4paper,top=3cm,bottom=2cm,left=3cm,right=3cm,marginparwidth=1.75cm]{geometry}

\newcommand{\appropto}{\mathrel{\vcenter{
  \offinterlineskip\halign{\hfil$##$\cr
    \propto\cr\noalign{\kern2pt}\sim\cr\noalign{\kern-2pt}}}}}

\usepackage{amsmath}
\usepackage{amssymb}
\usepackage{verbatim}
\usepackage{graphicx}
\usepackage{natbib}
\usepackage{caption}
\usepackage{subcaption}
\usepackage{floatrow}
\newfloatcommand{capbtabbox}{table}[][\FBwidth]

\usepackage{xcolor}
\usepackage{soul}

\newcommand{\alkcomment}[1]{\textcolor{purple}{\textit{[ALK: #1]}}}

\makeatletter
\def\BState{\State\hskip-\ALG@thistlm}
\makeatother

\usepackage{bbm}
\usepackage{algorithm}
\usepackage[noend]{algpseudocode}

\usepackage{authblk}

\title{Feedback alignment in deep convolutional networks}

\author[1,2,*]{Theodore H. Moskovitz}
\author[1]{Ashok Litwin-Kumar}
\author[1]{L.F. Abbott}

\affil[1]{\small Mortimer B. Zuckerman Mind, Brain and Behavior Institute, Department of Neuroscience, Columbia \protect\\ University, New York, NY}
\affil[2]{\small Department of Computer Science, Columbia University, New York, NY}
\affil[*]{\small \texttt{t.moskovitz@columbia.edu}}

\begin{document}
\maketitle


\begin{abstract}
\noindent Several recent studies have identified similarities between neural representations in biological networks and in deep artificial neural networks. This has led to renewed interest in developing analogies between the backpropagation learning algorithm used to train artificial networks and the synaptic plasticity rules operative in the brain. These efforts are challenged by biologically implausible features of backpropagation, one of which is a reliance on symmetric forward and backward synaptic weights. A number of methods have been proposed that do not rely on weight symmetry but, thus far, these have  failed to scale to deep convolutional networks and complex data. We identify principal obstacles to the scalability of such algorithms and introduce several techniques to mitigate them. We demonstrate that a modification of the \textit{feedback alignment} method that enforces a weaker form of weight symmetry, one that requires agreement of weight sign but not magnitude, can achieve performance competitive with backpropagation. Our results complement those of \citet{hinton_lilli_scale2018} and \citet{poggio_new2018} and suggest that mechanisms that promote alignment of feedforward and feedback weights are critical for learning in deep networks.
\end{abstract}

\section{Introduction}
While the hierarchical processing performed by deep neural networks is inspired by the brain, there are a number of fundamental differences between these artificial networks and their biological counterparts. In particular, the use of the backpropagation (BP) algorithm \citep{bprop_1986} to perform gradient descent, which is central to the optimization of artificial networks, requires several assumptions that are difficult to reconcile with biology. Objections include the separation of learning and inference into two separate phases and a requirement of symmetric synaptic connectivity between forward and backward paths through the network, an issue known as the \textit{weight transport problem} \citep{grossberg_1987}. While feedback connections are common, such symmetry has not yet been observed in the brain. 

\textit{Feedback alignment} (FA), a modification to BP in which this symmetry is broken by replacing the forward weights with randomized connections for the backward pass, avoids the weight transport problem \citep{Lillicrap2014}. While this method exhibits performance competitive with backpropagation in simple fully-connected networks \citep{Lillicrap2014,Nokland2016}, it has performed poorly when applied to deeper convolutional architectures and complex datasets \citep{Poggio2015,hinton_lilli_scale2018}. In this work, we explore the obstacles that hinder the performance of FA in deeper networks and present modifications that allow these methods to remain competitive with BP. We also experiment with the enforcement of fixed excitatory (E) and inhibitory (I) connectivity, and discuss its implications for learning in both the brain and artificial networks. 

\section{Related work}
The weight transport problem for artificial neural networks was identified early on \citep{grossberg_1987,Crick1989,Zipser1990}. While a number of potential solutions had been proposed previously \citep{Crick1989,brandt_lin,hinton_2003}, the FA method generated substantial interest because it required no assumptions on the structure of the feedback weights used to convey error signals, instead taking them to be fixed and random \citep{Lillicrap2014}. Initial work demonstrated that FA was competitive with BP on the MNIST handwritten digit dataset and a random input-output task in multilayer fully-connected networks. In these networks, it was observed that as training progresses, the angle between the BP gradient and the FA error signal converges from approximately orthogonal to roughly $45^{\circ}$, meaning that the FA weight updates are correlated with but not identical to those in BP. 

\citet{Poggio2015} applied FA to convolutional neural networks (CNNs), testing performance on a variety of tasks including visual and auditory classification. Without additional modifications, the performance of FA was substantially worse than BP in contrast to the earlier experiments on fully-connected networks. To achieve competitive performance, the authors made three modifications to the basic algorithm. The first of these is a technique termed uniform sign-concordant feedback (uSF), in which the feedback matrix $B$ is set to $B = \text{sign}(W)$, where $W$ represents the forward weights. The second is the addition of Batch Normalization \citep{ioffe_szegedy2015_bn}, and the third is a set of techniques termed Batch Manhattan in which the magnitude of the gradient is discarded. Batch Manhattan in particular was introduced to avoid vanishing/exploding gradients arising from the inherent weight asymmetry of FA. In this paper we demonstrate several alternative mechanisms for avoiding vanishing/exploding gradients without discarding the magnitude of the error signal.  

\citet{Nokland2016} introduced \textit{direct feedback alignment} (DFA), in which the output layer error signal is propagated directly to all upstream layers instead of through adjacent layers as in standard BP and FA. This idea was extended by \citep{Baldi2016} to include residual connections from downstream layers other than the output layer. Here, we further develop these ideas into new gradient methods, as well as demonstrate their success on deeper models than have been used previously. 

Recently, \citet{hinton_lilli_scale2018} presented results testing FA on CIFAR-10 and ImageNet architectures. It is important to note that, in in the interest of biological plausibility, their models eschewed the weight-sharing in convolutional layers that substantially improves generalization performance in most CNNs. Our models do take advantage of weight sharing to improve performance. A number of other approaches seek to further improve biological plausibility through the elimination of a distinct error signal altogether \citep{lecun_TP_1986, TP_Bengio14, Lee_2015_DTP} or the use of segregrated dendrites and continuous learning \citep{lillicrap_dendrites2016, bengio_dendrites2017}, but these approaches have also struggled when applied to deep networks and are outside the scope of our investigation.

During the final stages of the preparation of this manuscript, \citet{poggio_new2018} released results similar to ours on even deeper networks. Our results are consistent with theirs, and we extend the analysis by studying networks with fixed excitatory and inhibitory connections as well as several different modifications to FA. Together, these results provide strong evidence that sign-concordant feedback improves performance compared to random feedback.

\section{Feedback alignment in convolutional networks}
We begin by describing the implementation of FA in convolutional networks, which is similar to its implementation in fully-connected networks. For a convolutional layer $l$, the affine pre-activation $u^{l}$ and the $i$th nonlinear feature map activation $I^{l}_{i}$ are calculated as 
\begin{equation}
u^{l+1}_{i} = w^{l+1}_{i} \ast I^{l}_{i} + b^{l+1}\text{, } I^{l+1}_{i} = f(u^{l+1}_{i}),
\end{equation}
where $w_{i}$ is the $i$th kernel, $b$ is the bias term, $\ast$ denotes convolution, and $f$ is the activation function. Assuming a loss function $J$, the gradient as used in backpropagation at layer $l$ is 
\begin{equation}
\delta^{l}_{i} = \frac{\partial J}{\partial u^{l}_{i}} = (\overline{w}^{l+1}_{i} \ast \delta^{l+1}_{i}) \odot f'(u^{l}_{i}),
\end{equation}
where $\odot$ is component-wise multiplication and $\overline{w}$ denotes a rotation of $w$ by $180^\circ$, i.e. a \textit{flipped} kernel. FA replaces the feedforward kernel with a separate feedback matrix $B$ to produce the error signal 
\begin{equation}
\delta^{l}_{i} = \frac{\partial J}{\partial u^{l}_{i}} = (B^{l+1}_{i} \ast \delta^{l+1}_{i}) \odot f'(u^{l}_{i}).
\end{equation}
In both methods, the parameter updates are calculated similarly, with $\Delta W^{l} \propto \delta^{l}(x^{l-1})^{T}$ in fully-connected networks, and $\Delta W^{l} \propto \delta^{l}\ast \overline{I}^{l-1}$ in CNNs.

\begin{figure}[!t] 
\centering
\includegraphics[width=0.97\textwidth]{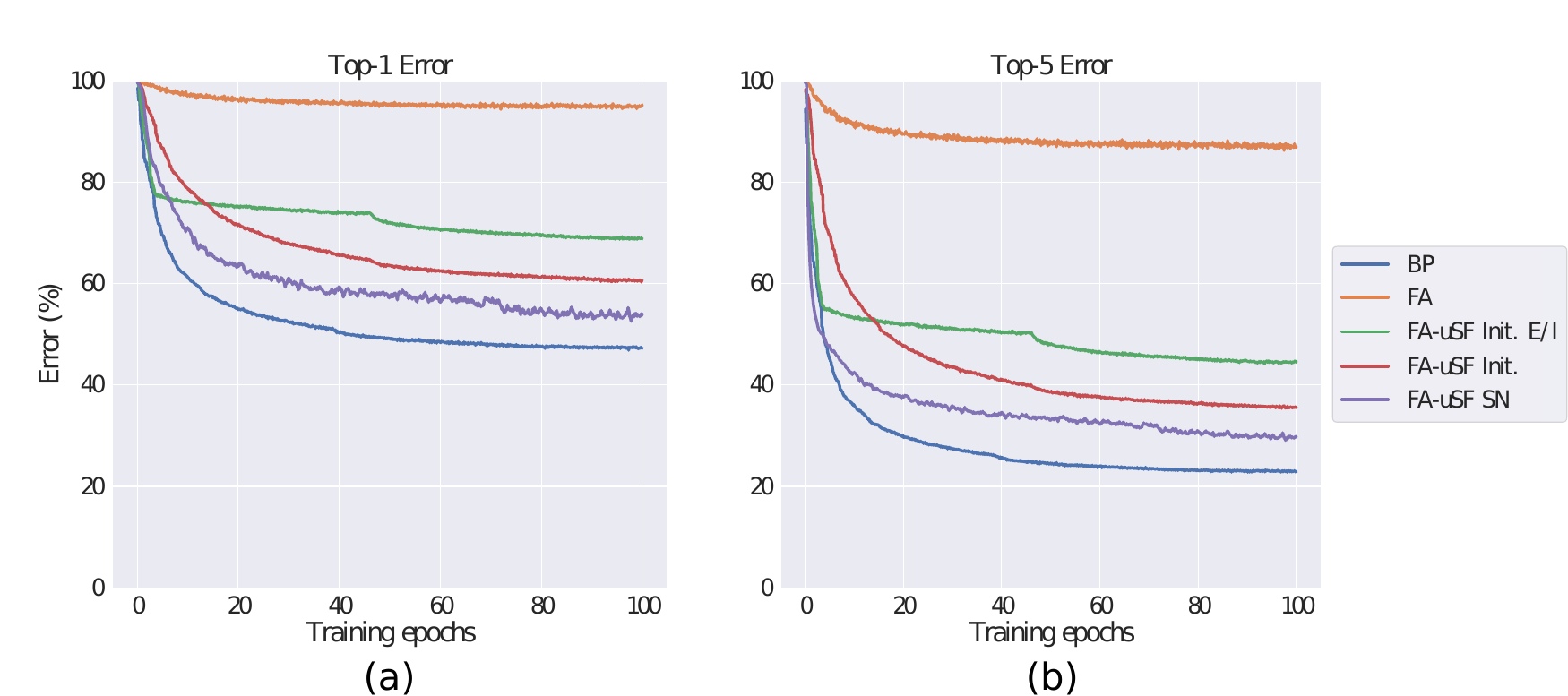}
\caption{Top-1 and Top-5 test errors on ImageNet for BP, FA without uSF or normalization (FA-no uSF), FA with uSF and the initialization method (FA-init), and FA with uSF and strict normalization (FA-strict-norm). See text for method details.}
\label{fig:err_curves}
\end{figure}

\subsection{Modifications to the FA algorithm} \label{sect:mods}
We now describe modifications to FA that improve performance by reducing the possibility of vanishing or exploding gradients and encouraging angular alignment between the feedforward and feedback weights. The reasons for their effectiveness are explored in detail in Section \ref{sect:analysis}. For uSF, the first feedback matrix setting we use is
\begin{equation}
    B^{l}_{t} = |B_{0}^{l}| \odot \text{sign}(W_{t}^{l}),
\end{equation}
where $|\cdot|$ is the element-wise absolute value function and $t$ denotes the training iteration. We call this technique, which uses no information about the magnitude of the current forward weights, the \textit{initialization} (Init.) method. One modification of the initialization method that we also tested was to enforce fixed excitatory/inhibitory connectivity, freezing the sign of the forward weights after a certain number of epochs of training. This imposes a new constraint on the network's synaptic connections, keeping them either excitatory (positive) or inhibitory (negative). Because under uSF the feedback weights equal the sign of the forward weights, this also results in a constant feedback matrix. We call this the \textit{excitatory/inhibitory} (E/I) method. We note that although this method prevents single synapses from changing sign during training, individual neurons may still form synapses of different signs, inconsistent with "Dale's Law."

The next setting we use, which incorporates the norm of the forward weights, is 
\begin{equation}
    B^{l}_{t} = ||W_{t}^{l}||_{2} \odot \frac{\text{sign}(W_{t}^{l})}{||\text{sign}(W_{t}^{l})||_{2}}.
\end{equation}
We call this the \textit{strict normalization} (SN) method. 
Note also that when uSF is used, some form of explicit normalization is required, as otherwise the magnitude of the feedback weights is $\pm 1$ regardless of the magnitude of the forward weights. 

\section{Experiments and classification results}

Each method described above was applied to deep CNNs tested on visual classification. Models were trained on the MNIST handwritten digits dataset, the CIFAR-10 image set, and the ImageNet dataset. On all three datasets, we implemented relatively simple baseline architectures following the basic paradigm established by LeCun et al. \citep{Lecun98gradient-basedlearning}, namely several convolutional layers each followed by a pooling layer, with one or several fully-connected layers on top. To test the performance of our proposed methods on models with increased depth, we also applied them to more complicated architectures with nine convolutional layers \citep{springenberg_allc2014} on CIFAR-10 and ImageNet.  The exact architecture details are presented in Supplementary Tables \ref{tab:archs} and \ref{tab:archs2}. In addition to FA, we trained models with direct feedback alignment (DFA) and a new method we call \textit{dense} feedback alignment (DenseFA), which adds residual feedback connections from every downstream layer. 
Unfortunately, the memory requirements of these methods precluded their use on the larger models we trained. To further investigate the relationship between FA and BP, we also investigated the performance of models that were trained with BP but with either added noise (BP + Noise) or with weight matrices that were forced to align with arbitrary random matrices (BP + Alignment). These techniques and their motivations are discussed in greater detail in Section \ref{sect:bp_mods}. On ImageNet, as a means of circumventing the need for normalization altogether, we also experimented with BP (BP Const.) and FA (FA-uSF Const.) models with weight norms constrained to be constant over training (see Section \ref{sect:norm_const} for details).

\begin{table}[!t]
\begin{center}
 \begin{tabular}{||c c c c c c||} 
 \hline
 Method & MNIST & CIFAR-10 1 & CIFAR-10 2 & ImageNet 1 & ImageNet 2\\ [0.5ex] 
 \hline\hline
 BP & 0.8 & 17.2 & 11.0 & 79.5 & 45.5 \\ 
 \hline
 BP $+$ Noise & 0.8 & 17.4 & 11.0 & 79.2 & 46.0 \\
 \hline 
 BP $+$ Alignment & 0.9 & 17.4 & 11.2 & 79.4 & 45.9 \\
 \hline 
 FA & 1.1 & 26.6 & 35.6 & 95.2 & 94.5 \\
 \hline 
 FA-uSF Init. E/I & 0.9 & 18.9 & 17.8 & 86.9 & 67.8 \\
 \hline 
 FA-uSF Init. & 0.7 & 17.6 & 13.1 & 79.6 & 60.1 \\
 \hline 
 FA-uSF SN & 0.7 & 17.7 & 12.6 & 78.9 & 54.4 \\
 \hline
 DFA & 1.0 & 28.6 & - & - & - \\
 \hline
 DenseFA & 0.7 & 16.9 & - & - & - \\
 \hline 
 BP Const. & - & - & - & - & 46.9 \\
 \hline 
 FA-uSF Const. & - & - & - & - & 51.2 \\
 \hline 
 FA-uSF Const. E/I & - & - & - & - & 66.1 \\
 \hline 
 \end{tabular}
 \caption{Top-1 Test Error (\%). The feedback alignment methods are competitive with, and in some cases exceed, backpropagation performance. `uSF' denotes the use of sign-concordant feedback. `Init.' denotes the initialization method, and `SN' denotes the strict normalization method. `E/I' denotes a model with frozen excitatory and inhibitory connections. These approaches are detailed in Section \ref{sect:mods}. `Const.' denotes a model for which the $L_{2}$-norms of the feedforward weights was fixed at initialization. This approach is described in Section \ref{sect:norm_const}. Model architectures are described in detail in Supplementary Section \ref{sect:archs}.}
 \label{tab:perf}
\end{center}
\end{table}

The MNIST model was trained for 25 epochs, the LeNet-style CIFAR-10 model was trained for 154 epochs, and the all-convolutional CIFAR-10 architecture was trained for 256 epochs. Both ImageNet models were trained for 100 epochs. In the E/I condition, we froze the signs of the weights (by clipping their values at just above or below zero) after 5\% of the training time (i.e., 5 epochs on ImageNet). All models were trained with the Adam optimizer \citep{adam_14}. Further training details can be found in Supplemental Section \ref{sect:supp_training_details}. Test results are summarized in Table \ref{tab:perf}, with ImageNet test error curves plotted in Figure \ref{fig:err_curves}. Our results improve on those reported by \cite{hinton_lilli_scale2018} and are consistent with those of \cite{poggio_new2018}.

\section{Scaling to deep networks} \label{sect:analysis}

We now describe in more detail the relevant considerations when extending FA to deeper networks and the modifications we made to improve performance.

\subsection{Vanishing and exploding activations and error signals}

\begin{figure}[!t] 
\centering
\includegraphics[width=0.95\textwidth]{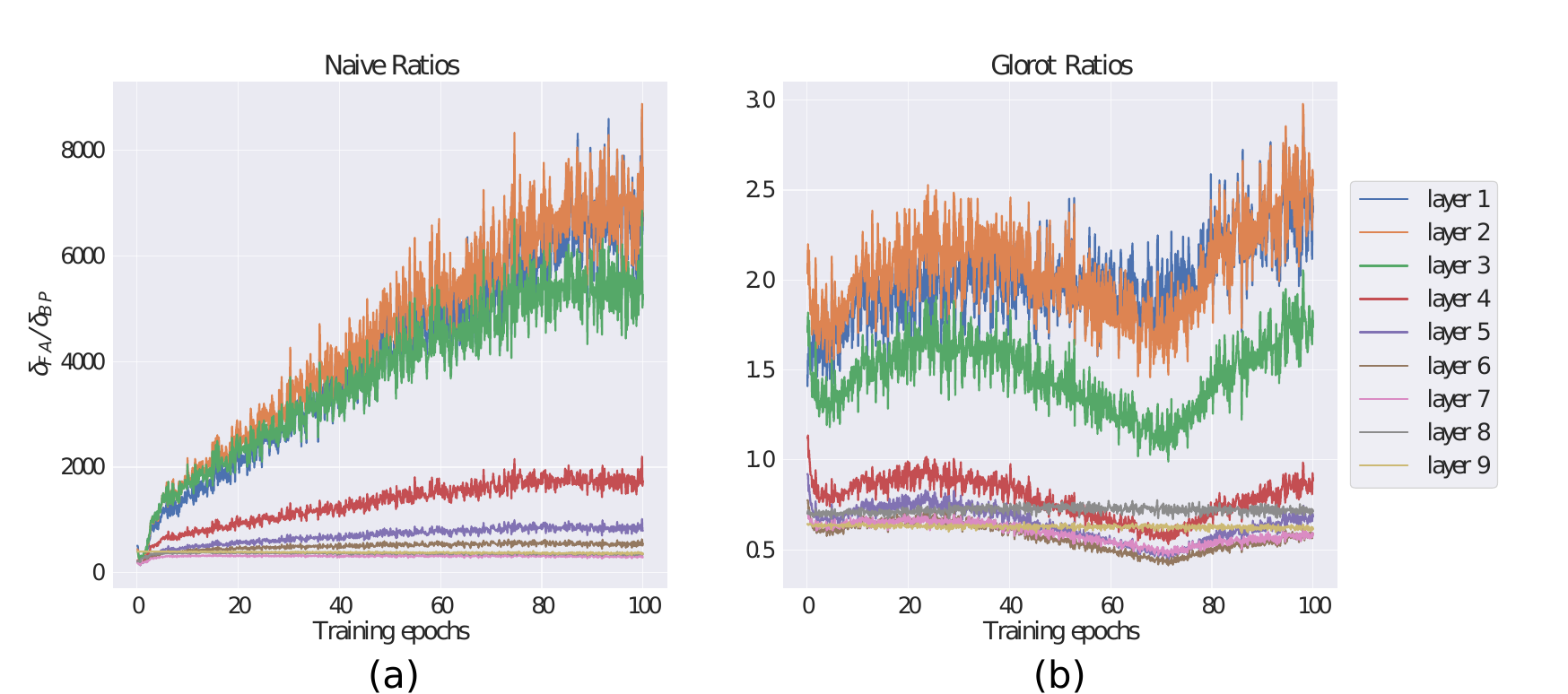}
\caption{The layer-wise gradient ratios between FA and BP for a na{\"i}ve initialization (a) and using a variance-preserving initialization such as the one devised by \cite{glorot_bengio_Xavier2010} (b). This demonstrates the importance of initialization in controlling for vanishing or exploding gradients as the depth of the network increases.}
\label{fig:inits}
\end{figure}

Consider a neural network with depth $L$ trained to minimize loss $J$. Assume for simplicity that the layers are sized equally so that each weight matrix is the same size and has a similar distribution of weights. The gradient with respect to the first layer activation $x^{1}$ is:
\begin{equation}
\frac{\partial J}{\partial x^{1}} = \frac{\partial J}{\partial x^{L}} \frac{\partial x^{L}}{\partial u^{L}} \frac{\partial u^{L}}{\partial x^{L-1}}
\frac{\partial x^{L-1}}{\partial u^{L-1}} \frac{\partial u^{L-1}}{\partial x^{L-2}} \dots \frac{\partial x^{2}}{\partial u^{2}} \frac{\partial u^{2}}{\partial x^{1}}.
\end{equation}
Observe that at each layer 
\begin{equation}
\frac{\partial u^{l}}{\partial x^{l-1}} = 
\begin{cases} 
      (W^{l})^{T} & \text{ if BP} \\
      B^{l} & \text{ if FA.} 
\end{cases}
\end{equation}
If the weights are assumed to be independent, then
\begin{equation}
\frac{||\nabla_{x^{1}}^{BP}J ||_{2}}{||\nabla_{x^{1}}^{FA}J)||_{2}} \appropto \prod _{l=1}^L \frac{(||W^l)^T||_2}{||B^l||_2},
\end{equation}
where $||\cdot||_{2}$ denotes the Euclidean norm. Because $||W^l||_{2}$ changes over the course of training and $||B^l||_{2}$ remains fixed, a network trained with feedback alignment runs a risk of experiencing vanishing or exploding gradients even if BP does not, depending on the ratio $||B^l||_{2}/||W^l||_{2}$. Moreover, this problem is exponential in the depth of the network (Figure \ref{fig:inits}a).


There are two possible families of approaches to solving this issue. First, careful initialization of the forward and backward weights can dramatically reduce fluctuations in the scale of activations and error signals between layers. Initialization methods designed to control variance from layer to layer as a means of effectively training deep networks are common \citep{glorot_bengio_Xavier2010, orthog_init2018}. One example is the initialization strategy introduced by \citep{glorot_bengio_Xavier2010} in which weights are initialized with a variance of $1 / [\frac{1}{2}(n_{\rm in}+n_{\rm out})]$,  with $n_{\rm in}$ the number of input connections from the previous layer to neurons of the subsequent layer and $n_{\rm out}$ the number of output connections. In BP, this method controls the variance of both the forward activations and the backward gradients by averaging the number of incoming and outgoing connections. However, in FA the forward and backward passes are decoupled, and the most effective initialization is therefore to set the variance of the forward weights as $1/n_{\rm in}$ and the variance of the fixed, backward weights as $1/n_{\rm out}$. Note that $n_{\rm out}$ is the number of incoming connections that a layer receives during the backward pass. While we found this method to be effective (Figure \ref{fig:inits}b), depth remains a challenge, as the distribution of the forward weights can drift during training. With careful initialization, the scale of the drift is constant to within an order of magnitude, but there is still an adverse effect on performance. 


The second family of approaches instead explicitly manages the sizes of the weights. For example, \citet{Poggio2015} introduced a parameter update rule termed \textit{Batch Manhattan}, which discards the magnitude of the error signal completely. That is, whereas the gradient descent weight update is proportional to $\frac{\partial J}{\partial W}$, Batch Manhattan sets it proportional to $\text{sign}\left(\frac{\partial J}{\partial W} \right)$. In discarding the magnitude of the gradient, this method ignores the impact that the size of this signal has on the effective learning rate of the network. The use of an adaptive, per-parameter optimization algorithm such as Adam \citep{adam_14} can ameliorate this to a degree, but even in this case we found a reduction in performance. We found that a more effective method was either to constrain the norms of the feedback matrices to be equal to that of their feedforward counterparts, as in SN, or to simply scale by the initialized feedback weights, as in the initialization method.

\subsection{Gradient signal alignment}
\begin{figure}[!t] 
\centering
\includegraphics[width=1.0\textwidth]{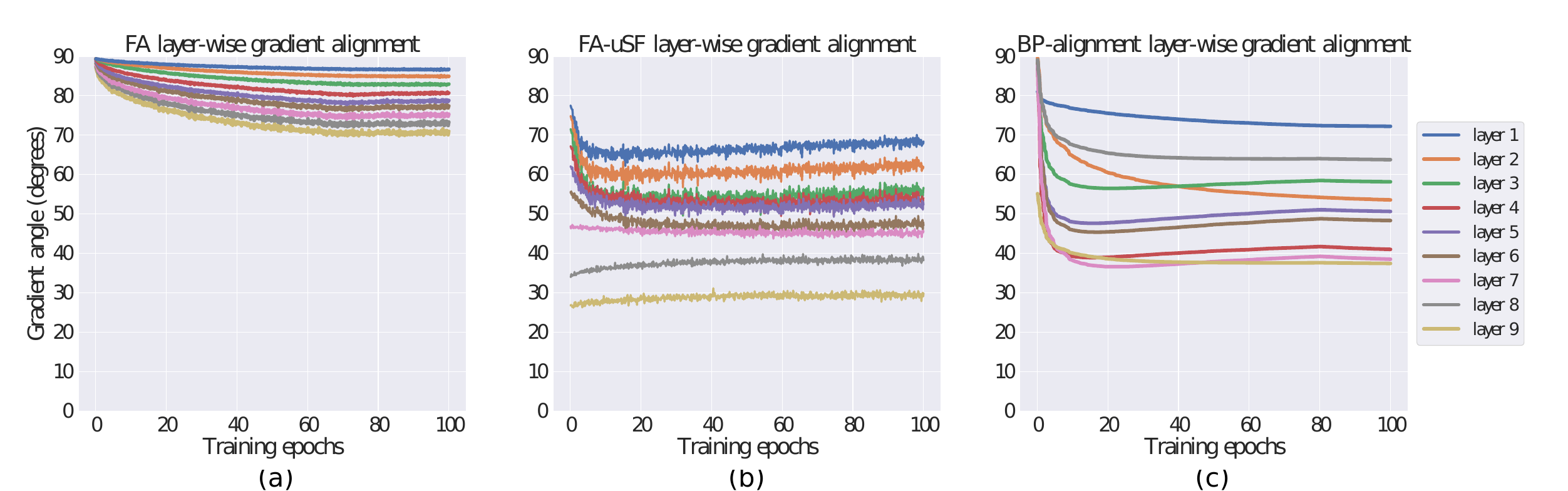}
\caption{The layer-wise angular alignment between the gradient computed with BP and with (a) FA, (b) FA with uSF, and (c) BP with an alignment constraint. Although the angles are comparable, constrained BP still slightly outperforms FA (Table \ref{tab:perf}). These results demonstrate that viable solutions can be found without strictly following the gradient.}
\label{fig:alignment}
\end{figure}

When investigating FA in shallow feedforward networks, \citep{Lillicrap2014} observed that as training progressed, the signal calculated in FA and the gradient used by BP are roughly orthogonal at initialization but that the angle between them converges to approximately $45^{\circ}$ by the end of training. While we also observed this convergence in shallow networks, deeper networks exhibited less alignment. More precisely, while we found that the angle converged to some degree for the topmost layers, this was not the case for the lower layers (Figure \ref{fig:alignment}a). In order to maintain alignment, we used the uSF method introduced by \citep{Poggio2015}. When uSF is applied, the angle quickly drops below $90^{\circ}$, but then rises slightly and levels off (Figure \ref{fig:alignment}b). This is likely because at the beginning of training, depending on the initialization method, forward weight values are more tightly distributed than at the end of training.


\subsection{Modifications to BP} \label{sect:bp_mods}
Because of the relationship between alignment and performance, we investigated models that were trained with BP but with modifications that mimic features of FA. These included either adding noise to the error signal or forcing the forward weights to align with an unrelated matrix (see Table \ref{tab:perf} for results). In the first condition, we added noise drawn from a normal distribution centered at zero with variance approximately equal to that of the BP gradients. The angle between the noisy learning signals and that of the true gradients remained at roughly $45^\circ$ throughout training, comparable to the alignment achieved by FA with uSF. This did not substantially reduce performance (Table \ref{tab:perf}), indicating that the reduction in performance for FA cannot be accounted for by gradient noise that is centered on the BP gradient. 

In the second condition, we applied an $L_{2}$ penalty to the difference between the model weights $\theta = \{w^{1},...,w^{L}\}$ and a random set of target matrices $\Lambda = \{v^{1},...,v^{L}\}$ (where $L$ is the depth of the network) in addition to the cross entropy loss:
\begin{equation}
J(y,\hat{y},\theta,\Lambda) = -\sum_{k=1}^{K}y\log\hat{y} + \lambda \sum_{l=1}^{L}\sum_{i,j} (w_{ij}^{l} - v_{ij}^{l})^{2},
\end{equation}
where $y$ and $\hat{y}$ are the true and predicted logits, respectively, $K$ is the number of label classes, and $\lambda$ is the regularization weight. This penalty forces the forward weights to align with a set of fixed random matrices, as in FA, but without using the separate matrix for error propagation (Figure \ref{fig:alignment}). We found that setting $\lambda=0.001$ was effective at producing an alignment comparable to that of FA. Performance did not noticeably suffer as a result of either of these changes. This indicates that the constraint of aligning with an arbitrary matrix does not in itself limit the performance of deep networks.

\subsection{Constraining the magnitude of the weights} \label{sect:norm_const}

\begin{figure}[!t]
  \centering
  \begin{minipage}[c]{0.38\textwidth}
    \centering
    \begin{tabular}{||c c c||} 
 \hline
 Method & Top-1 & Top-5 \\ [0.5ex] 
 \hline\hline
 BP & 46.9 & 22.6 \\ 
 \hline
 FA-uSF & 51.2 & 26.6 \\
 \hline 
 FA-uSF EI & 66.1 & 35.2 \\
 \hline 
 \end{tabular}
  \end{minipage}
  \begin{minipage}[c]{0.58\textwidth}
    \includegraphics[width=\textwidth]{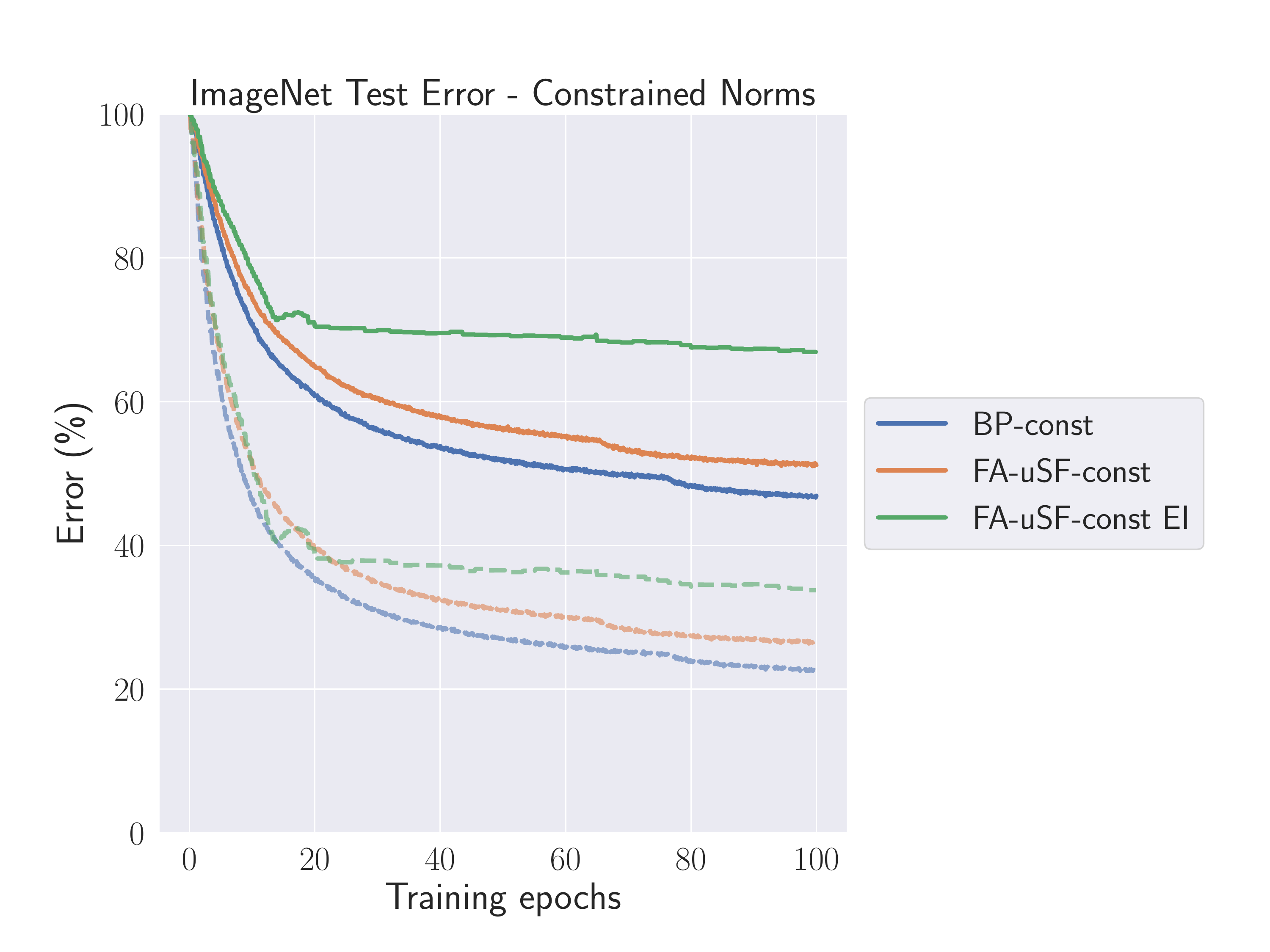}
  \end{minipage}
  \caption{Top-1 (solid) and Top-5 (dotted) ImageNet test results and error curves for models trained with the $L_{2}$-norms of their forward weights fixed at the beginning training. This condition eliminates the need for the training-time normalization methods, and further narrows the gap between BP and FA. }\label{fig:results_const}
\end{figure}

While the various normalization methods we have introduced are effective at reducing the effect that the changing magnitude of the feedforward weights has on FA over time, to completely circumvent this issue, we trained several models on ImageNet in which the norm of the forward weights was adjusted after each training iteration. Specifically, the weights were initialized normally, and after each iteration $t$ the weights at each layer $l$ were scaled as follows:
\begin{equation}
    w_{t}^{l} = ||w_{0}^{l}||_{2} \frac{w_{t}^{l}}{||w_{t}^{l}||_{2}},
\end{equation}
where $||w_{0}^{l}||_{2}$ is the $L_{2}$-norm of the initial weights. Using a variance-preserving initialization method (Figure \ref{fig:inits}), this condition allows us to evaluate the effectiveness of FA without the need for additional mechanisms to normalize the backward weights as the forward weights change. The results demonstrate that constraining the norm in this way does not substantially affect the performance of BP, while resulting in improved performance using FA with $51.2\%$ Top-1 error, further narrowing the gap between the two methods (Table \ref{tab:perf}, Figure \ref{fig:results_const}). 

\subsection{Effect of constrained weight signs}
\begin{figure}[!t] 
\centering
\includegraphics[width=0.6\textwidth]{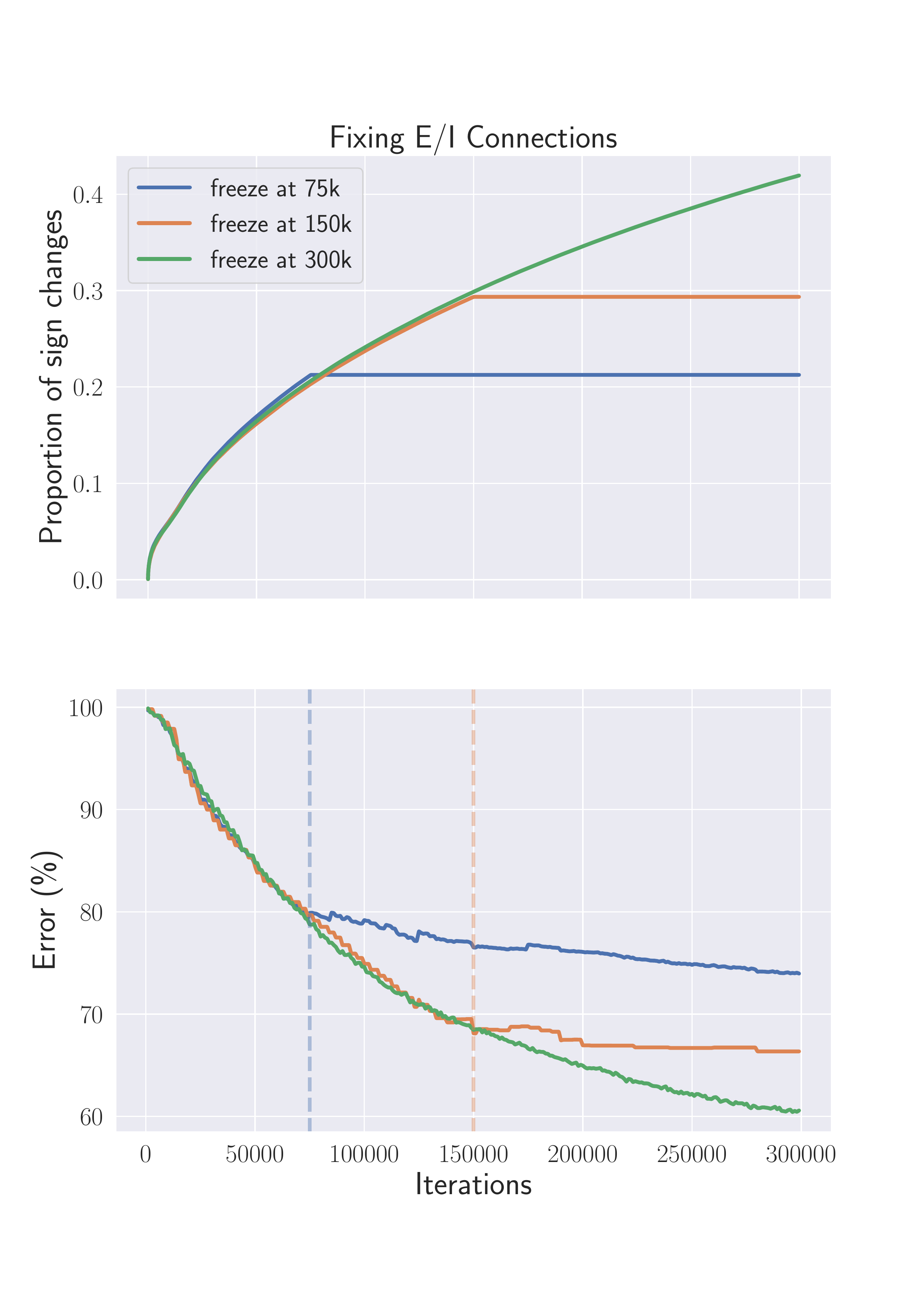}
\caption{Allowing weights to change sign has a dramatic effect on network performance. The top panel examines the effect of enforcing fixed weight signs at various points in the initial stages of training on ImageNet test performance for models trained with FA-uSF and fixed weight norms. The bottom panel displays the proportion of weights that had changed signs since initialization for each condition. We can see that, somewhat surprisingly, weights continue to flip sign at a high rate, and that allowing this mobility has a direct impact on network performance.}
\label{fig:freeze_EI}
\end{figure}

Our results in Table \ref{tab:perf} and Figure \ref{fig:results_const} demonstrate that freezing weight signs has a significant negative impact on learning. If weight signs stabilize over the course of training, constraining them to remain either excitatory or inhibitory after this stabilization should have a minimal effect on performance. However, if weights do not stabilize, enforcing sign constraints may negatively impact performance even if the constraints are enforced after a large number of training epochs. To assess the stability of weight signs over training, we tracked the cumulative fraction of weights that exhibited a change in sign as a function of training iterations (Figure \ref{fig:freeze_EI}). Even after 300,000 training mini-batches into, many weight signs were still changing, indicating that excitatory or inhibitory identities are not fixed even late in training. Accordingly, constraining weight signs dramatically reduces the speed of learning, even if the constraints are applied late in training. It is unclear what impact the choice of initialization has on this phenomenon. However, these results appear to be a substantial challenge for biologically plausible models in which not only the signs of single synapses, but of all synapses formed by an individual neuron, are fixed in accordance with Dale's Law.

\section{Discussion}
Taken together, these results extend previous work and demonstrate that modified FA learning algorithms can perform with accuracy competitive with BP even for deep CNNs. Necessary modifications include controlling the magnitude of the error signal, which can be accomplished using the normalization methods we investigated, and encouraging alignment, which can be accomplished with sign-concordant feedback.  These conclusions are consistent with those of \citet{poggio_new2018}. In our simulations, we found that employing these methods along with fixed weight norms allow our model to reach an ImageNet Top-1 error of $51.2\%$, which is competitive with the same architecture trained with BP. A topic of experimental interest is identifying biological mechanisms with roles analogous to these methods.

Homeostatic mechanisms regulating feedback connections could play a role in normalizing forward and backward synaptic weights  \citep{turrigiano_homeostatic_2004}. We studied a variety of mechanisms to accomplish this normalization, from explicitly scaling the feedback weights to match the feedforward weights in norm to constraining the feedforward weights to have an unchanging norm. We found that these methods improved upon methods that ignore gradient norms \citep{Poggio2015}. Fixing the feedforward weight norm (Figure \ref{fig:results_const}) led to the best results, suggesting that regulating the forward pass and normalizing the backward weights appropriately may be sufficient to achieve high performance. We focused only on instantaneous normalization, but in the future, it would be valuable to experiment with scaling the weights with a delay that reflects the time constant of homeostatic regulation. 

Sign-concordant feedback was also crucial to performance, consistent with the idea that ``vanilla'' feedback alignment does not scale to deep networks \citep{hinton_lilli_scale2018,poggio_new2018}. An attractive possibility is that the segregation of biological neurons into excitatory and inhibitory subtypes permits cell-type-specific wiring that promotes sign-concordant feedback. However, the adverse effects of enforcing fixed E/I connections (Figure \ref{fig:freeze_EI}) represent a significant obstacle to biological realism. Our experiments show not only that unconstrained neurons continue to switch between excitatory and inhibitory modes throughout learning, but also that the freedom to do so is directly connected with improved task performance. Developing further understanding of this phenomenon is important for understanding not only the brain, but also the learning process in deep networks. 

While our results and those of \citet{poggio_new2018} represent a step toward understanding the plausibility of such an algorithm, many questions remain, including how performance is affected in networks without weight sharing \citep{hinton_lilli_scale2018} and how to incorporate continuous learning without separate forward and backward passes \citep{lillicrap_dendrites2016, bengio_dendrites2017}. Further experimental and computational studies are needed to develop clearer analogies between biological and artificial neural network learning. 


\newpage 
\bibliographystyle{abbrvnat}
\bibliography{iclr2019_conference}

Research was supported by a Burroughs-Wellcome Award (A.L.-K.) and by NSF NeuroNex Award DBI-1707398 and the Gatsby Charitable Foundation.

\newpage 
\section{Supplemental information}
\subsection{Training Information} \label{sect:supp_training_details}
In the MNIST network, the initial learning rate was set to $\eta=0.0001$, and was reduced to $1\times 10^{-5}$ after $20$ epochs. A stepped learning rate decay was also used in the CIFAR-10 models. In the LeNet-style model, the initial learning rate was set to $\eta=0.0005$, and decayed by a factor of $0.8$ every $30$ epochs. In the all-convolutional model, the initial learning rate was set to $\eta=0.05$, and multiplied by a factor of $0.1$ every $120$ epochs. A weight decay with strength $\lambda = 0.001$ was also added to all layers. The batch size was set to $n=50$ for the MNIST network and $n=128$ for the CIFAR-10 models. Dropout \citep{hinton_dropout14} was applied with a drop probability of $0.5$ after the densely-connected layer in the MNIST network (layer 5). In the all-convolutional architecture, dropout was applied after the downsampling layers (3 and 6) with a drop probability of $0.5$, as well after the input layer with a drop probability of $0.2$.

\subsection{Model architectures} \label{sect:archs}
\begin {table}[H] 
\begin{center}
 \begin{tabular}{||c c c c||} 
 \hline
 Layer & MNIST & CIFAR-10 Model 1 & CIFAR-10 Model 2 \\ [0.5ex] 
 \hline\hline
 Input & $28 \times 28 \times 1$ & $24 \times 24 \times 3$* & $32 \times 32 \times 3$\\ 
 \hline
 1 & $5\times 5$ conv. $32$ ReLU & $5\times 5$ conv. $64$ ReLU & $3\times 3$ conv. $96$ ReLU \\ 
 \hline
 2 & $2\times 2$ max-pool stride 2&  $2\times 2$ max-pool stride 2 & $3\times 3$ conv. $96$ ReLU \\
 \hline
 3 & $5\times 5$ conv. $64$ ReLU  & $5\times 5$ conv. $64$ ReLU & $3\times 3$ conv. $96$ ReLU stride $2$\\
 \hline
 4 & $2\times 2$ max-pool stride 2 & $2\times 2$ max-pool stride 2 & $3\times 3$ conv. $192$ ReLU \\
 \hline
 5 & 1024 dense ReLU & 384 dense ReLU & $3\times 3$ conv. $192$ ReLU \\
 \hline
 6 & 10-way softmax & 192 dense ReLU & $3\times 3$ conv. $192$ ReLU stride $2$ \\
 \hline
 7 & - & 10-way softmax & $3\times 3$ conv. $192$ ReLU \\
 \hline
 8 & - & - & $1\times 1$ conv. $192$ ReLU \\
 \hline
 9 & - & - & $1\times 1$ conv. $10$ ReLU \\
 \hline
 10 & - & - & global average pooling ($8 \times 8$ dim.)  \\
 \hline
 11 & - & - & 10-way softmax  \\
 \hline
 \end{tabular}
 \caption{Model architectures. *CIFAR-10 images were cropped as part of data augmentation to increase the size of the training set.}
 \label{tab:archs}
\end{center}
\end{table}

\begin {table}[H] 
\begin{center}
 \begin{tabular}{||c c c||} 
 \hline
 Layer & ImageNet Model 1 & ImageNet Model 2 \\ [0.5ex] 
 \hline\hline
 Input & $299 \times 299 \times 3$ & $299 \times 299 \times 3$\\ 
 \hline
 1 & $9\times 9$ conv. $192$ ReLU stride 2 & $9\times 9$ conv. $192$ ReLU stride 4\\ 
 \hline
 2 &  $2\times 2$ max-pool stride 2 & $3\times 3$ conv. $192$ ReLU stride 2\\
 \hline
 3 & $5\times 5$ conv. $192$ ReLU stride 2 & $3\times 3$ conv. $192$ ReLU stride 3\\
 \hline
 4 & $2\times 2$ max-pool stride 2 & $5\times 5$ conv. $256$ ReLU stride 2\\
 \hline
 5 & 512 dense ReLU & $3\times 3$ conv. $256$ ReLU stride 1\\
 \hline
 6 & 512 dense ReLU & $3\times 3$ conv. $256$ ReLU stride 2 \\
 \hline
 7 & 1000-way softmax & $3\times 3$ conv. $512$ ReLU stride 1 \\
 \hline
 8 & - & $1\times 1$ conv. $512$ ReLU stride 1\\
 \hline
 9 & - & $1\times 1$ conv. $1000$ ReLU stride 1\\
 \hline
 10 & - & global average pooling ($4 \times 4$ dim.)  \\
 \hline
 11 & - & 1000-way softmax  \\
 \hline
 \end{tabular}
 \caption{ImageNet architectures. These are scaled variations of the two CIFAR-10 architectures above.}
 \label{tab:archs2}
\end{center}
\end{table}

\end{document}